\documentclass[runningheads]{llncs}
\usepackage[T1]{fontenc}
\usepackage{graphicx}
\usepackage[hidelinks]{hyperref}
\usepackage{multirow}
\usepackage{amsmath}
\usepackage{amsfonts} 
\usepackage{booktabs}
\usepackage{adjustbox}
\usepackage{tabularx,ragged2e}
\usepackage{amsfonts}       
\usepackage{nicefrac}       
\usepackage{microtype}      
\usepackage{xcolor}         
\usepackage{todonotes}
\usepackage{placeins}
\usepackage{notoccite}
\usepackage{orcidlink}
\usepackage{comment}
\usepackage{subcaption}

\usepackage{hyperref}

\begin{document}
\title{ODExAI: A Comprehensive Object Detection Explainable AI Evaluation}
\titlerunning{ODExAI: Object Detection Explainable AI Evaluation}
%
\author{Loc Phuc Truong Nguyen\inst{1}\orcidlink{0009-0003-4254-0750} \and
Hung Truong Thanh Nguyen\inst{2}\orcidlink{0000−0002−6750−9536} \and
Hung Cao\inst{2}\orcidlink{0000−0002−0788−4377}}
\authorrunning{Loc Nguyen et al.}
%
\institute{Friedrich-Alexander-Universität Erlangen-Nürnberg, 91054 Erlangen, Germany \and
University of New Brunswick, Fredericton E3B 5A3, Canada
\email{loc.pt.nguyen@fau.de, \{hung.ntt,hcao3\}@unb.ca}}
\maketitle              
\begin{abstract}
Explainable Artificial Intelligence (XAI) techniques for interpreting object detection models remain in an early stage, with no established standards for systematic evaluation. This absence of consensus hinders both the comparative analysis of methods and the informed selection of suitable approaches. To address this gap, we introduce the \textit{Object Detection Explainable AI Evaluation} (ODExAI), a comprehensive framework designed to assess XAI methods in object detection based on three core dimensions: localization accuracy, faithfulness to model behavior, and computational complexity. We benchmark a set of XAI methods across two widely used object detectors (YOLOX and Faster R-CNN) and standard datasets (MS-COCO and PASCAL VOC). Empirical results demonstrate that region-based methods (e.g., D-CLOSE) achieve strong localization (PG = 88.49\%) and high model faithfulness (OA = 0.863), though with substantial computational overhead (Time = 71.42s). On the other hand, CAM-based methods (e.g., G-CAME) achieve superior localization (PG = 96.13\%) and significantly lower runtime (Time = 0.54s), but at the expense of reduced faithfulness (OA = 0.549). These findings demonstrate critical trade-offs among existing XAI approaches and reinforce the need for task-specific evaluation when deploying them in object detection pipelines. Our implementation and evaluation benchmarks are publicly available at: \url{https://github.com/Analytics-Everywhere-Lab/odexai}.

\end{abstract}

\section{Introduction}
Deep neural networks (DNNs) have greatly advanced object detection, but their opaque decision-making limits interpretability \cite{ijcai2024p1025,NGUYEN2025102782,FindingtheRightXAIMethodAGuidefortheEvaluationandRankingofExplainableAIMethodsinClimateScience}. This reduces trust and restricts deployment in sensitive domains such as healthcare \cite{nguyen2023towards,WOOD2022102391,jpm11111213,nguyen2025human} and security \cite{10444383,10723939,Natarajan2024}, where models must offer not only high accuracy but also transparent, fair, and domain-aligned reasoning \cite{Teng2022,10.3389/frobt.2024.1444763,YANG202229}. Explainable AI (XAI) addresses this need by making detection decisions more interpretable.

XAI has emerged as a distinct research area, driven by advances in deep learning, DARPA-funded initiatives \cite{Gunning_Aha_2019}, and the recognition of a right to explanation in the GDPR \cite{hoofnagle2019european}, EU AI Act \cite{neuwirth2022eu}. In recent years, a wide range of XAI methods for object detectors has been proposed. While this reflects growing interest in the field, it also presents several challenges  \cite{ijcai2023p747}. The number of available methods is increasing rapidly, and many offer only minor variations, such as Spatial Sensitive Grad-CAM \cite{9897350} and Spatial Sensitive Grad-CAM++\cite{Yamauchi_2024_CVPR}, or produce similar outputs despite differing implementations, as seen with D-RISE \cite{Petsiuk_2021_CVPR} and ODAM \cite{10478163}. As discussed by several works \cite{10.1145/3583558,ijcai2024p1025,NGUYEN2025102782,FindingtheRightXAIMethodAGuidefortheEvaluationandRankingofExplainableAIMethodsinClimateScience,10444383}, explanation quality must be assessed across several dimensions, such as faithfulness to the model’s reasoning, conciseness, fairness, and alignment with domain knowledge. Yet two critical questions remain open: \textit{how to reliably select an appropriate XAI method}, and \textit{how to benchmark it in a standardized manner} that ensures objective evaluation and supports generalizable findings.

To address these challenges, we introduce the \textit{Object Detection Explainable AI Evaluation} (ODExAI), a structured approach for evaluating and comparing XAI methods for object detection. Recent work has focused on developing metrics to benchmark interpretability techniques across settings. These metrics enable quantitative assessment of localization, faithfulness, and complexity, offering insight into each method’s strengths, limitations, and suitability. The main contributions of this work are:
\begin{enumerate}
    \item We propose ODExAI Framework grounded in three key interpretability dimensions: faithfulness, localization, and complexity. Each dimension is supported by a set of quantitative metrics, enabling systematic and task-specific assessment.

    \item We conduct a toy experiment to illustrate the use of ODExAI in evaluating three novel XAI methods: D-CLOSE \cite{pmlr-v222-truong24a}, G-CAME \cite{nguyen2024efficientconciseexplanationsobject}, and D-RISE \cite{Petsiuk_2021_CVPR}, across two widely used detectors, YOLOX \cite{ge2021yoloxexceedingyoloseries} and Faster R-CNN \cite{ren2016fasterrcnnrealtimeobject}, using the MS COCO \cite{10.1007/978-3-319-10602-1_48} and PASCAL VOC \cite{pascal-voc-2012} datasets.

    \item We conduct a comprehensive analysis of the trade-offs between XAI methods, highlighting their performance across different interpretability dimensions. Based on the evaluation outcomes, we provide practical guidelines for selecting suitable XAI methods according to task-specific requirements.
\end{enumerate}

\section{Background}
This section reviews object detection models and the XAI methods applied to interpret their outputs. It also outlines recent progress and key challenges, particularly the lack of a unified and standardized evaluation framework for explanation techniques.

\subsection{Object Detection Models}
DNNs have significantly advanced object detection through various techniques, each offering distinct advantages. Single-stage detectors, such as the YOLO \cite{redmon2016lookonceunifiedrealtime} family (e.g., YOLOv3 \cite{redmon2018yolov3incrementalimprovement}, YOLOv4 \cite{bochkovskiy2020yolov4optimalspeedaccuracy}, and YOLOX \cite{ge2021yoloxexceedingyoloseries}) and SSD\cite{Liu_2016}, are optimized for real-time detection, offering a balance between speed and accuracy. Two-stage detectors, known for their high precision, include Faster R-CNN \cite{NIPS2015_14bfa6bb}, Mask R-CNN \cite{8237584} (which extends detection to instance segmentation), and Cascade R-CNN \cite{8578742}, which refines results through multi-stage processes. Anchor-free detectors simplify detection by directly predicting object centers or corners, with notable examples such as CornerNet \cite{law2019cornernetdetectingobjectspaired}, CenterNet \cite{9010985}, and FCOS \cite{9010746} (Fully Convolutional One-Stage Object Detection). Transformer-based detectors, including DETR \cite{9010746} (DEtection TRansformer), Deformable DETR \cite{carion2020endtoendobjectdetectiontransformers} (which accelerates convergence), and Swin Transformer \cite{9710580} (known for hierarchical feature extraction), leverage attention mechanisms to improve performance. 

\subsection{Explainable AI for Object Detection}
While XAI techniques are widely used in classification and segmentation, their application to object detection remains limited due to constraints in flexibility, efficiency, and adaptability \cite{8689279}. As noted in \cite{nguyen2024efficientconciseexplanationsobject}, these methods fall into two main categories: region-based and CAM-based saliency approaches. Region-based methods assess the importance of image regions by masking inputs and observing changes in output. Examples include LIME \cite{10.1145/2939672.2939778}, RISE \cite{petsiuk2018riserandomizedinputsampling}, and object detection variants such as SODEx \cite{make3030033}, D-RISE \cite{Petsiuk_2021_CVPR}, and D-CLOSE \cite{pmlr-v222-truong24a}, which enhance precision through task-specific adaptations. These approaches are model-agnostic and intuitive. In contrast, CAM-based methods like Grad-CAM \cite{Selvaraju_2017_ICCV} use feature maps and partial derivatives to generate saliency maps, offering efficiency but often yielding less informative results for detection tasks. Designed primarily for classification, they lack dedicated adaptations for detection. The recently proposed G-CAME \cite{nguyen2024efficientconciseexplanationsobject} addresses these gaps by providing stable, efficient explanations across both one-stage and two-stage detectors.

\subsection{Development and Gaps in XAI for Object Detection}
XAI is essential for object detection for understanding how models assign class labels and bounding boxes from visual features. It helps verify whether predictions rely on relevant regions, supports error analysis, and enhances transparency. In healthcare, XAI has been applied to thyroid nodule detection \cite{nguyen2023towards}, tumor localization \cite{jpm11111213}, and abnormality identification \cite{WOOD2022102391}. In autonomous driving, it aids in diagnosing detection failures \cite{s24216776} and assessing robustness under varying weather conditions \cite{abc}. In environmental monitoring, it explains outputs in deforestation tracking \cite{BUCHELT2024121530} and ecological change analysis \cite{inproceedings}. In security, it promotes fairness in detecting individuals with disabilities \cite{10444383}, while in agriculture, it supports interpretation in crop health monitoring \cite{Natarajan2024} and pest detection \cite{10723939}.

Although XAI methods are broadly applicable, their evaluation remains inconsistent and lacks standardization \cite{ijcai2023p747,arya2019explanationdoesfitall}. This stems from subjective assessment criteria and methodological diversity. In many cases, formal procedures are absent, leading to ad hoc strategies tailored to specific models, datasets, or use cases \cite{app12199423}. As a result, findings are often difficult to compare or reproduce, limiting progress toward reliable and generalizable explanations. Several initiatives have attempted to address this gap. Quantus \cite{hedstrom2023quantus}, an open-source toolkit, offers computational metrics for evaluating explanation properties. Other studies seek to clarify evaluation processes \cite{ijcai2023p747,10.1145/3583558} or provide practical guidelines for designing evaluation protocols \cite{10.1145/3387166,arya2019explanationdoesfitall}. However, current approaches lack consensus, and no unified evaluation framework exists for XAI for object detection  \cite{nguyen2023towards,MORADI2024109183,10297629}.

\section{The proposed ODExAI Framework}
The absence of a standardized evaluation framework in XAI has led to the use of ad hoc metrics, often tailored to specific methods. Although these metrics cover diverse evaluation criteria, their proliferation introduces uncertainty about task relevance, hindering reproducibility and generalization. To address this, ODExAI defines three core evaluation properties for object detection, each illustrated by schematic diagrams (\autoref{loco}–\autoref{comp}). Metrics are grouped by property based on shared objectives, enabling a more structured, consistent, and interpretable evaluation process.

\subsection{Framework Overview}
\begin{figure}[ht!]
    \centering
    \includegraphics[width=0.8\linewidth]{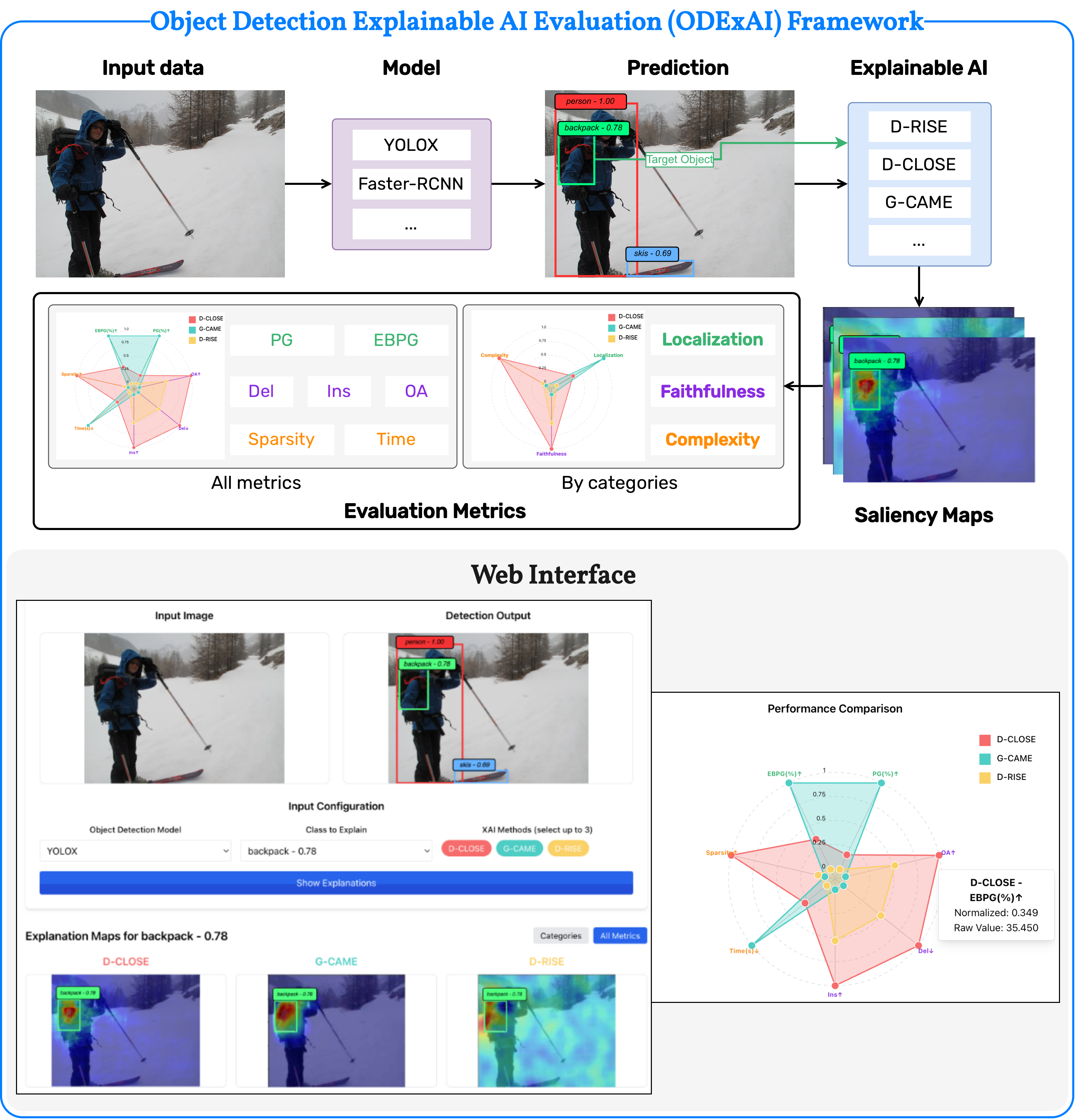}
    \caption{Overview of the architecture and web interface of our proposed Object Detection Explainable AI Evaluation (ODExAI).}
    \label{frame}
\end{figure}

Our framework consists of two main components: Saliency Map Extraction with XAI and Explanation Evaluation. The overall procedure is illustrated in \autoref{frame}. A detailed description of each component is provided below:
\begin{enumerate}
    \item \textbf{Saliency Map Extraction with XAI}: This part focuses on generating saliency maps using XAI methods. Users begin by uploading an image and selecting the task, with object detection models assigned, e.g., Faster R-CNN \cite{ren2016fasterrcnnrealtimeobject} and YOLOX \cite{ge2021yoloxexceedingyoloseries}. After prediction, the user selects the target object and an explanation method, e.g., D-CLOSE \cite{pmlr-v222-truong24a}, G-CAME \cite{nguyen2024efficientconciseexplanationsobject}, and D-RISE \cite{Petsiuk_2021_CVPR}, to generate the saliency map.
    \item \textbf{Explanation Evaluation}: This part provides an overview of each explanation method across three dimensions: localization, faithfulness, and complexity. Once the explanation is generated, users can initiate the evaluation, which is visualized as a spider plot. Each dimension includes multiple metrics, with corresponding numerical values that quantify the method’s performance.
\end{enumerate}

\subsection{Localization Metrics}
\begin{figure}[ht]
    \centering
    \includegraphics[width=0.8\linewidth]{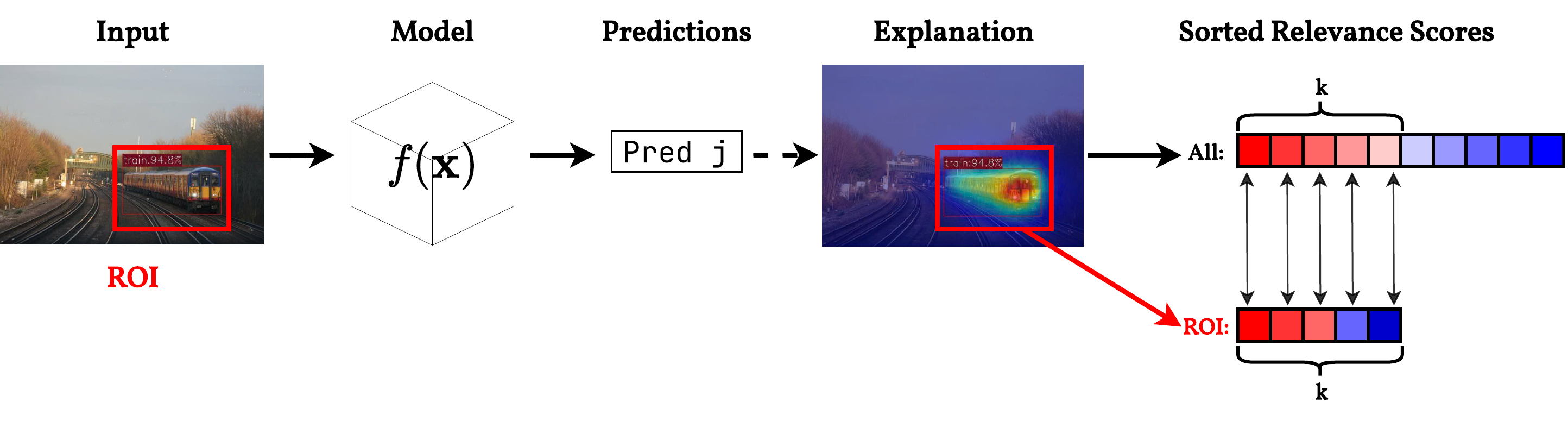}
    \caption{Demonstration of the localization property using region-specific relevance comparisons. Localization measures how well an explanation aligns with the spatial region that is most responsible for the model’s prediction. To evaluate this, a region of interest (ROI) is predefined in the input image (red box), marking the area where the most relevant evidence is expected to reside. In this example, the target is the object “train”. After model inference, an explanation map is generated, and pixel-wise relevance scores are computed. The $k$ relevance scores within the ROI are then compared with the top-$k$ scores from the full explanation map. A method with strong localization is expected to assign the highest relevance to pixels inside the ROI.}
    \label{loco}
\end{figure}

Localization quality is measured by how well an explanation aligns with a user-defined region of interest (ROI) \cite{FindingtheRightXAIMethodAGuidefortheEvaluationandRankingofExplainableAIMethodsinClimateScience}, as shown in Figure~\ref{loco}. This involves comparing the most relevant pixels in the explanation map to labeled regions such as bounding boxes or segmentation masks. Since the ROI is assumed to be critical to the model’s decision \cite{FindingtheRightXAIMethodAGuidefortheEvaluationandRankingofExplainableAIMethodsinClimateScience,arras2022clevr,zhang2018top}, localization is considered high when relevance is concentrated within it. We quantify this using the Pointing Game (PG) \cite{zhang2018top} and Energy-based Pointing Game (EBPG) \cite{wang2020score} metrics.

\subsubsection{Pointing Game (PG)}
\cite{zhang2018top} evaluates the spatial precision of saliency maps by measuring their ability to localize target object categories. For each saliency map, the metric identifies the point of maximum saliency and checks if it falls within the ground truth annotation of the target object. A hit is recorded if the point lies within the annotated region; otherwise, it is counted as a miss. The localization accuracy is defined as:
\begin{align}
    \text{PG} = \frac{\text{\# Hits}}{\text{\# Hits}+\text{\# Misses}}
\end{align}
where $\text{\# Hits}$ and $\text{\# Misses}$ denote the total number of correct and incorrect localizations, respectively. The overall performance is calculated by averaging the accuracy across all categories. A high PG score for an XAI method indicates its ability to generate accurate and interpretable saliency maps. 

\subsubsection{Energy-Based Pointing Game (EBPG)}
\cite{wang2020score} evaluates the spatial accuracy of saliency maps by analyzing how much of the map’s energy (activation) falls within the annotated bounding box of the target object. Unlike the traditional PG, which focuses on a single maximum point, this metric considers the overall distribution of energy within the relevant regions. The localization accuracy is defined as:
\begin{align}
\text{EBPG} = \frac{\sum L^c_{(i,j) \in \text{bbox}}}{\sum L^c_{(i,j) \in \text{bbox}} + \sum L^c_{(i,j) \not\in \text{bbox}}}
\end{align}
where $L^c_{(i,j)}$ represents the saliency map intensity at pixel $(i,j)$, and $\text{bbox}$ refers to the bounding box region of the target object. The overall performance is calculated by averaging the accuracy across all categories. A high EBPG value indicates that the XAI-generated saliency maps can effectively localize the target region and suppress noise outside the bounding box. 

\subsection{Faithfulness Metrics}
\begin{figure}[t]
    \centering
    \includegraphics[width=0.8\linewidth]{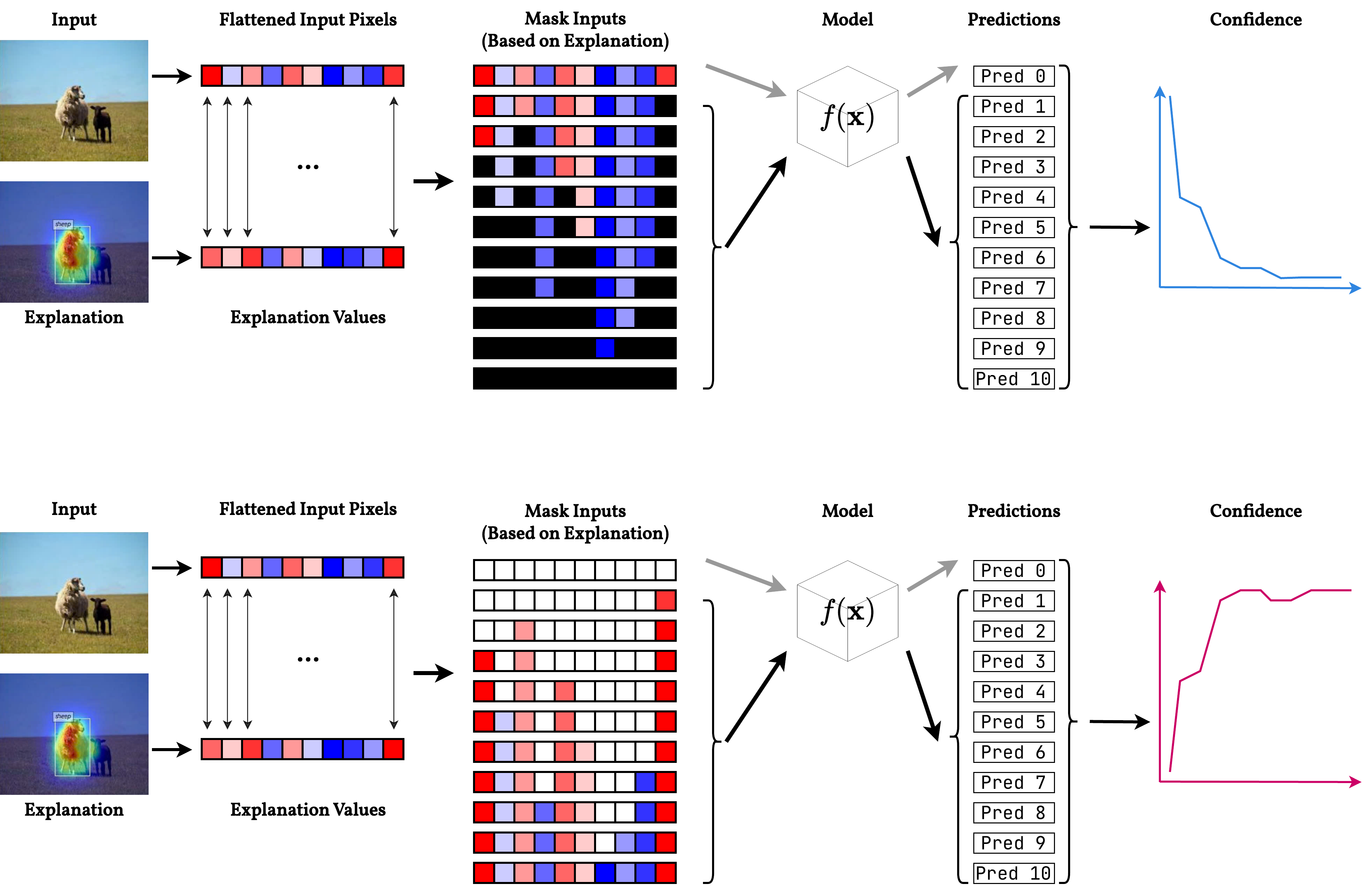}
    \caption{Illustration of the faithfulness property using progressive perturbation. Faithfulness evaluates how the most relevant pixels affect the model’s prediction. 
    Explanation scores are first ranked to identify the most relevant features, highlighted in red. Their corresponding positions are then mapped onto the flattened input, as indicated by the arrows. In the top panel, pixels are progressively removed (masked in black), while in the bottom panel, they are progressively added to a blank input. At each step, the modified input is passed through the model, and the prediction confidence is recorded. A faithful explanation produces a steep confidence drop when highly relevant pixels are removed and a sharp increase when they are added, whereas perturbing less relevant regions yields smaller changes.}
    \label{faith}
\end{figure}

Faithfulness measures how changes to highly relevant input features affect the model’s prediction \cite{pmlr-v222-truong24a}, as shown in Figure~\ref{faith}. It is evaluated by progressively inserting or deleting pixels with high relevance scores and comparing the resulting predictions to the original. Since relevance reflects each feature’s contribution, significant prediction shifts are expected when important features are modified. We quantify faithfulness using the Deletion and Insertion \cite{petsiuk2018riserandomizedinputsampling} and Over-All (OA) \cite{zhang2021group} metrics.

\subsubsection{Deletion (Del) and Insertion (Ins)} 
The Deletion \cite{petsiuk2018riserandomizedinputsampling} metric evaluates the effect of removing key features by tracking the decline in predicted class probability as important pixels, identified by the saliency map, are progressively removed. A steep probability drop, quantified by a low area under the curve (AUC), denotes a strong explanation. In contrast, the Insertion \cite{petsiuk2018riserandomizedinputsampling} metric measures the probability increase when pixels are gradually introduced, with a higher AUC indicating better explanations. 

\subsubsection{Over-All (OA)} \cite{zhang2021group} metric is defined as $\text{AUC}(\text{Insertion}) - \text{AUC}(\text{Deletion})$, provides a comprehensive evaluation of saliency map quality by considering both its positive and negative effects on the model’s predictions. A higher score indicates the saliency map effectively highlights critical regions (high Insertion) while minimizing the influence of irrelevant ones (low Deletion).

\subsection{Complexity Metrics}
\begin{figure}[t!]
    \centering
    \includegraphics[width=0.8\linewidth]{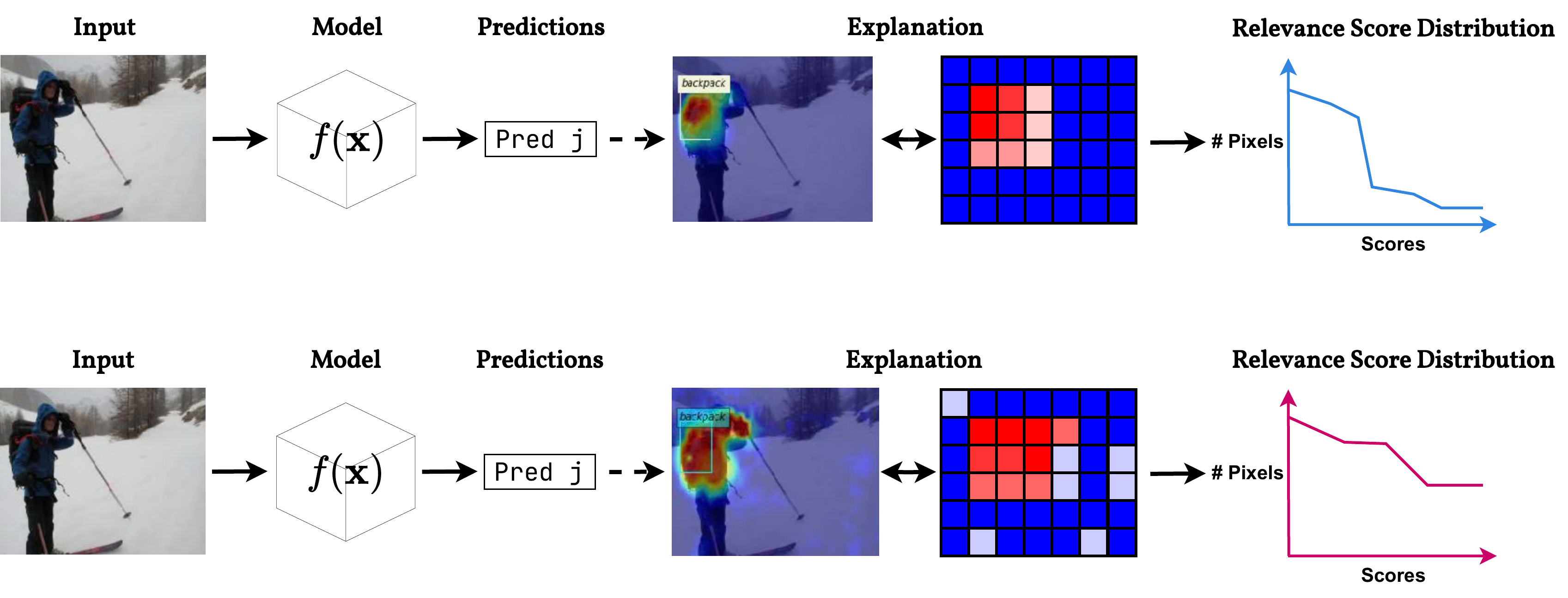}
    \caption{Visualization of the complexity property via relevance score distributions. Complexity measures how concentrated or dispersed an explanation is over the input space. After inference, relevance scores are assigned to all input pixels and sorted to generate a distribution curve. In the top panel, low complexity is illustrated by a steep curve where few pixels carry most of the relevance. In contrast, the bottom panel shows high complexity, with scores distributed more uniformly, resulting in a flatter curve. Low-complexity explanations are desirable because they highlight key regions, reduce the effective input space, and focus the analysis on essential model components, leading to more efficient and interpretable outputs.}
    \label{comp}
\end{figure}

Complexity reflects the conciseness of an explanation, ideally highlighting a small set of key features \cite{FindingtheRightXAIMethodAGuidefortheEvaluationandRankingofExplainableAIMethodsinClimateScience}. Concise explanations are easier to interpret and tend to convey more useful information with less noise \cite{FindingtheRightXAIMethodAGuidefortheEvaluationandRankingofExplainableAIMethodsinClimateScience,10.1007/978-3-031-09037-0_8}. We also incorporate computation time into the complexity measure, as high runtime can limit the practicality of XAI methods in real-time or large-scale systems \cite{KAKOGEORGIOU2021102520}. Complexity is quantified using the Sparsity metric \cite{10.1007/978-3-031-09037-0_8} and by recording the runtime of each method. The overall concept is illustrated in \autoref{comp}.

\subsubsection{Sparsity}
\cite{10.1007/978-3-031-09037-0_8} metric measures how concentrated or dispersed the scores in a saliency map are, reflecting the focus of attention on specific regions. It is defined as the ratio of the maximum score ($S_\text{max}$) to the mean score ($S_\text{mean}$):
\begin{align}
    \text{Sparsity} = \frac{S_\text{max}}{S_\text{mean}}
\end{align}
It should be noted that all saliency map scores are min–max normalized beforehand to ensure consistency across methods. A high sparsity value corresponds to a high $S_\text{max}/S_\text{mean}$ ratio, i.e., a low mean score $S_\text{mean}$, indicating that the map's focused areas are narrow and well-defined. 

\subsubsection{Computation Time}
records the time taken to compute explanations for a given model input. It is typically reported as the average runtime per instance and depends on factors such as model complexity, input size, and explanation method. Lower values indicate higher efficiency, while higher values reflect greater computational cost.
\section{Demonstrative Example}
This section presents a toy experiment to demonstrate the utility of the ODExAI framework. We first introduce the evaluated XAI methods, benchmark datasets, and object detection architectures, followed by the experiment and a detailed report of the results.

\subsection{Benchmark Datasets and Object Detection Models}
We evaluate XAI methods on two standard object detection benchmarks: \textbf{MS COCO 2017} \cite{10.1007/978-3-319-10602-1_48}, containing 123,287 images across 80 object categories, and \textbf{PASCAL VOC 2012} \cite{pascal-voc-2012}, with 11,530 images spanning 20 classes. These datasets are selected for their diversity in scene complexity and annotation quality. To ensure architectural variety, we employ \textbf{Faster R-CNN} \cite{ren2016fasterrcnnrealtimeobject}, a two-stage detector known for high accuracy, and \textbf{YOLOX} \cite{ge2021yoloxexceedingyoloseries}, a single-stage, anchor-free model optimized for real-time performance.

\subsection{XAI Methods for Object Detection}
In this work, we focus on XAI methods from the region-based and CAM-based categories, as defined in \cite{nguyen2024efficientconciseexplanationsobject} for object detection models. A brief description of the selected methods is given as follows:
\begin{enumerate}
    \item \textbf{D-CLOSE} \cite{pmlr-v222-truong24a} is a region-based XAI method that generates saliency maps by segmenting the input image at multiple levels and combining their features to enhance explanation quality. Random masks are used to perturb the input, and the similarity scores between the masked outputs and the target prediction are calculated to weigh the importance of each region. The saliency maps are normalized using a density map to reduce noise, and multi-scale feature fusion aggregates fine-grained to coarse-grained features. 

    \item \textbf{D-RISE} \cite{Petsiuk_2021_CVPR} is a region-based XAI method designed to generate saliency maps for object detectors by jointly addressing localization and classification. It applies randomized binary masks to the input and evaluates their impact on detection outcomes. The importance of each masked region is estimated using a similarity metric that combines Intersection over Union (IoU) for spatial alignment, cosine similarity for classification consistency, and objectness scores. The final saliency map is obtained by aggregating the masks weighted by their contributions.

    \item \textbf{G-CAME} \cite{nguyen2024efficientconciseexplanationsobject} improves CAM-based explanation methods for object detection by integrating a Gaussian kernel for saliency map refinement. Specifically, it combines gradient-based localization with Gaussian weighting, effectively suppressing background noise and enhancing the detection of small objects. The method consists of four steps: selection of target layers, gradient-based object localization, feature map weighting, and Gaussian masking. 
\end{enumerate}

\subsection{Results}

\begin{figure}[b!]
    \centering
    \begin{subfigure}[b]{0.35\textwidth}
        \centering
        \includegraphics[width=\textwidth]{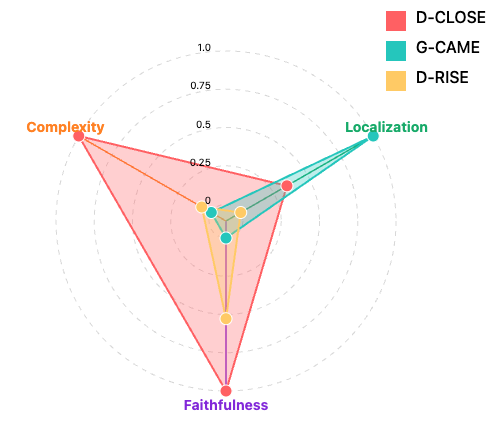}
        \caption{Metrics by categories}
        \label{fig:subfig1}
    \end{subfigure}
    \begin{subfigure}[b]{0.35\textwidth}
        \centering
        \includegraphics[width=\textwidth]{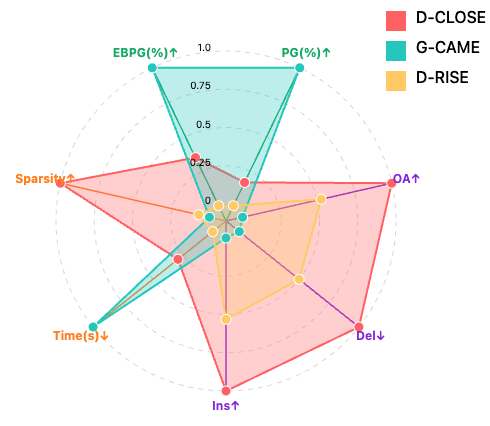}
        \caption{All metrics}
        \label{fig:subfig2}
    \end{subfigure}
    \caption{A visualization of the ODExAI framework for quantitative metrics of evaluated XAI methods for YOLOX on the MS COCO dataset: (a) representative metrics by categories (OA for Faithfulness, EBPG for Localization, and Sparsity for Complexity) and (b) all metrics.}
    \label{spider}
\end{figure}

We evaluate the selected XAI methods for object detection using YOLOX and Faster R-CNN on the MS COCO and PASCAL VOC datasets across three core criteria: faithfulness, localization, and complexity, each measured using dedicated quantitative metrics. All experiments are conducted under identical settings, with 2000 perturbed masks for both D-CLOSE and D-RISE. The evaluation is run on a system with an Nvidia Tesla T4 GPU and 24GB of RAM. Results in \autoref{restable} and \autoref{spider} show that ODExAI reliably measures explanation quality and informs method selection for specific detection tasks.

\begin{table}[b!]
\caption{Evaluation results from ODExAI.}
\label{restable}
\begin{center}
\begin{adjustbox}{width=1\textwidth}
\begin{tabular}{c ccc ccc ccc ccc}
\toprule
\textbf{Dataset} & \multicolumn{6}{c}{\textbf{MS-COCO}} & \multicolumn{6}{c}{\textbf{PASCAL VOC}} \\
\cmidrule[0.5pt]{2-13}
Model & \multicolumn{3}{c}{YOLOX} & \multicolumn{3}{c}{Faster R-CNN} & \multicolumn{3}{c}{YOLOX} & \multicolumn{3}{c}{Faster R-CNN} \\
Method & D-CLOSE & G-CAME & D-RISE & D-CLOSE & G-CAME & D-RISE & D-CLOSE & G-CAME & D-RISE & D-CLOSE & G-CAME & D-RISE \\
\midrule
\multicolumn{1}{c|}{Ins$\uparrow$} & \textbf{0.908} & 0.703 & \multicolumn{1}{c|}{0.812} & \textbf{0.912} & 0.718 & \multicolumn{1}{c||}{0.867} & \textbf{0.804} & 0.512 & \multicolumn{1}{c|}{0.775} & \textbf{0.826} & 0.534 & 0.783 \\
\multicolumn{1}{c|}{Del$\downarrow$} & \textbf{0.027} & 0.059 & \multicolumn{1}{c|}{0.043} & \textbf{0.049} & 0.169 & \multicolumn{1}{c||}{0.152} & 0.128 & 0.183 & \multicolumn{1}{c|}{\textbf{0.103}} & \textbf{0.171} & 0.284 & 0.206 \\
\multicolumn{1}{c|}{OA$\uparrow$} & \textbf{0.881} & 0.644 & \multicolumn{1}{c|}{0.769} & \textbf{0.863} & 0.549 & \multicolumn{1}{c||}{0.715} & \textbf{0.676} & 0.329 & \multicolumn{1}{c|}{0.672} & \textbf{0.655} & 0.250 & 0.577 \\
\midrule
\multicolumn{1}{c|}{PG(\%)$\uparrow$} & 87.86 & \textbf{94.31} & \multicolumn{1}{c|}{86.55} & 88.49 & \textbf{96.13} & \multicolumn{1}{c||}{84.12} & 81.48  & \textbf{93.96} & \multicolumn{1}{c|}{79.01} & 82.94 & \textbf{95.13}  & 80.11  \\
\multicolumn{1}{c|}{EBPG(\%)$\uparrow$} & 35.45 & \textbf{67.16} & \multicolumn{1}{c|}{18.47} & 36.13 & \textbf{70.11} & \multicolumn{1}{c||}{20.74} & 29.23 & \textbf{58.39} & \multicolumn{1}{c|}{28.11} & 31.84 & \textbf{57.84} & 29.94 \\
\midrule
\multicolumn{1}{c|}{Sparsity$\uparrow$} & \textbf{25.02} & 2.84 & \multicolumn{1}{c|}{4.43} & \textbf{26.13} & 4.95 & \multicolumn{1}{c||}{5.91} & \textbf{26.84} & 3.81 & \multicolumn{1}{c|}{12.45} & \textbf{27.12} & 4.91 & 6.21 \\
\multicolumn{1}{c|}{Time(s)$\downarrow$} & 70.12 & \textbf{0.43} & \multicolumn{1}{c|}{98.67} & 71.42 & \textbf{0.54} & \multicolumn{1}{c||}{99.11} & 70.94 & \textbf{0.48} & \multicolumn{1}{c|}{99.19} & 72.94 & \textbf{0.63} & 99.98 \\
\bottomrule
\end{tabular}
\end{adjustbox}
\end{center}
\end{table}

\subsubsection{Faithfulness Results}
Across both datasets and models, D-CLOSE consistently outperformed in faithfulness metrics. For example, on MS-COCO with YOLOX, it achieved the highest Insertion (0.908) and OA (0.881), with the lowest Deletion (0.027). This trend held across all settings, with D-CLOSE achieving the highest OA on three out of four model–dataset combinations. These results indicate that region-based explanations generated by D-CLOSE are tightly aligned with the model’s true decision process, highlighting its high faithfulness.

\subsubsection{Localization Results}
In contrast, G-CAME excelled in the localization dimension, achieving the highest EBPG and PG scores across all configurations. For instance, on YOLOX with MS COCO, it reached PG = 94.31\% and EBPG = 67.16\%, significantly outperforming D-CLOSE and D-RISE. This consistent strength in localization suggests that CAM-based methods are highly effective at identifying relevant object regions, even when their explanations do not fully capture the model’s internal reasoning.

\subsubsection{Complexity Results}
G-CAME was also the most computationally efficient, with runtimes under 1s across all settings. For instance, on MS-COCO with YOLOX, it required only 0.43s, compared to 70.12s for D-CLOSE and 98.67s for D-RISE. However, its Sparsity was the lowest, indicating more diffuse explanations. In contrast, D-CLOSE produced the most concise maps but at a significantly higher cost.

\section{Discussion}
This section discusses the trade-offs between XAI methods revealed by ODExAI, offers guidance on selecting suitable approaches, acknowledges current limitations, and outlines directions for future development.

\subsection{Trade-off Analysis and Suggestions for Method Selection}
The evaluation reveals clear trade-offs among XAI methods for object detection, showing that no single approach satisfies all interpretability requirements \cite{padmanabhan2023dextdetectorexplanationtoolkit,Molnar2022}. Region-based methods like D-CLOSE demonstrate strong faithfulness and produce sparse saliency maps due to their perturbation-based design: they systematically mask input regions and measure changes in model predictions, directly attributing importance to causally influential areas \cite{pmlr-v222-truong24a,nguyen2024efficientconciseexplanationsobject}. This yields explanations closely aligned with the model’s decision process, as confirmed by faithfulness metrics, and results in concise, object-focused maps with minimal noise \cite{pmlr-v222-truong24a}. However, the repeated inference across multiple segmentation levels imposes significant computational overhead, limiting practicality for real-time or resource-constrained applications \cite{pmlr-v222-truong24a,nguyen2024efficientconciseexplanationsobject}.

In contrast, CAM-based methods such as G-CAME achieve superior localization and low runtime by using internal feature maps and gradient signals from specific layers to generate saliency maps \cite{nguyen2024efficientconciseexplanationsobject}. On the one hand, selecting appropriate layers for explanation requires specialized knowledge from users \cite{famiglini2024evidence,NGUYEN2025102782}. On the other hand, this design enables rapid computation and produces highly localized explanations, as reflected in strong PG and EBPG scores. However, because they rely on aggregated activations rather than direct input perturbations, these methods may miss the exact features responsible for decisions, leading to lower faithfulness \cite{nguyen2024efficientconciseexplanationsobject,mumuni2025explainableartificialintelligencexai}.

D-RISE, while not leading in any single metric, maintains consistently moderate performance across all evaluation dimensions, making it a robust option for ensemble-based detection or sensitivity analysis, where explanation diversity is valuable \cite{pmlr-v222-truong24a,nguyen2024efficientconciseexplanationsobject,Petsiuk_2021_CVPR}. These results demonstrate ODExAI’s effectiveness in highlighting method-specific trade-offs and guiding the choice of XAI techniques based on task-specific interpretability goals.

\subsection{Limitations and Future Development}
Despite promising results, the current framework presents several limitations. First, the number of included XAI methods and evaluation metrics remains limited, which may restrict generalizability. Future work will expand the framework to incorporate additional methods and metrics, aiming for broader coverage and the development of more comprehensive evaluation tools. Second, some dimensions mix higher-better and lower-better metrics, which may introduce inconsistency. A possible solution is to assign weights to account for these directional differences. Third, the framework lacks a proper normalization scheme, which can be misleading when metrics differ in scale or unit. A promising approach is the skill score proposed in \cite{FindingtheRightXAIMethod}, which compares an explanation method’s performance relative to a reference baseline designed to violate the assumptions of each evaluation property. The skill score quantifies how much an explanation improves over this worst-case scenario.

\section{Conclusion}
In this work, we introduce the \textit{Object Detection Explainable AI Evaluation} (ODExAI), a structured system for assessing XAI methods for object detection. ODExAI supports the selection of explanation techniques based on three core dimensions: localization accuracy, faithfulness to model predictions, and computational complexity. To validate the framework, we evaluate several XAI methods on two widely used object detectors and standard benchmark datasets. Results show that region-based methods perform well in both localization and faithfulness but incur higher computational cost. CAM-based methods, by contrast, are more efficient and provide good localization but are less faithful. These findings reveal key trade-offs and underscore the need for task-specific evaluation when selecting XAI methods. Future work will expand the framework with additional methods and metrics, and establish consistent, interpretable evaluation procedures for each dimension.

\section*{Acknowledgment}
This work is supported by the UNB-FCS Startup Fund (22–23 START UP/H CAO).


\bibliographystyle{splncs04}
\bibliography{references}

\begin{thebibliography}{10}
\providecommand{\url}[1]{\texttt{#1}}
\providecommand{\urlprefix}{URL }
\providecommand{\doi}[1]{https://doi.org/#1}

\bibitem{arras2022clevr}
Arras, L., et~al.: Clevr-xai: A benchmark dataset for the ground truth evaluation of neural network explanations. Information Fusion  \textbf{81},  14--40 (2022)

\bibitem{arya2019explanationdoesfitall}
Arya, V., et~al.: One explanation does not fit all: A toolkit and taxonomy of ai explainability techniques (2019)

\bibitem{bochkovskiy2020yolov4optimalspeedaccuracy}
Bochkovskiy, A., et~al.: Yolov4: Optimal speed and accuracy of object detection (2020)

\bibitem{FindingtheRightXAIMethodAGuidefortheEvaluationandRankingofExplainableAIMethodsinClimateScience}
Bommer, P.L., et~al.: Finding the right xai method—a guide for the evaluation and ranking of explainable ai methods in climate science. Artificial Intelligence for the Earth Systems  \textbf{3}(3),  e230074 (2024)

\bibitem{FindingtheRightXAIMethod}
Bommer, P.L., et~al.: Finding the right xai method—a guide for the evaluation and ranking of explainable ai methods in climate science. Artificial Intelligence for the Earth Systems  \textbf{3}(3),  e230074 (2024)

\bibitem{BUCHELT2024121530}
Buchelt, A., et~al.: Exploring artificial intelligence for applications of drones in forest ecology and management. Forest Ecology and Management  \textbf{551},  121530 (2024)

\bibitem{10723939}
C, B., et~al.: A novel approach for jute pest detection using improved vgg-19 and xai. In: 2024 15th International Conference on Computing Communication and Networking Technologies (ICCCNT). pp.~1--5 (2024)

\bibitem{8578742}
Cai, Z., Vasconcelos, N.: Cascade r-cnn: Delving into high quality object detection. In: 2018 IEEE/CVF Conference on Computer Vision and Pattern Recognition. pp. 6154--6162 (2018)

\bibitem{carion2020endtoendobjectdetectiontransformers}
Carion, N., et~al.: End-to-end object detection with transformers (2020)

\bibitem{9010985}
Duan, K., et~al.: Centernet: Keypoint triplets for object detection. In: 2019 IEEE/CVF International Conference on Computer Vision (ICCV). pp. 6568--6577 (2019)

\bibitem{10.3389/frobt.2024.1444763}
Ennab, M., Mcheick, H.: Enhancing interpretability and accuracy of ai models in healthcare: a comprehensive review on challenges and future directions. Frontiers in Robotics and AI  \textbf{11} (2024)

\bibitem{jpm11111213}
Esmaeili, M., et~al.: Explainable artificial intelligence for human-machine interaction in brain tumor localization. Journal of Personalized Medicine  \textbf{11}(11) (2021)

\bibitem{pascal-voc-2012}
Everingham, M., et~al.: The {PASCAL} {V}isual {O}bject {C}lasses {C}hallenge 2012 {(VOC2012)} {R}esults

\bibitem{famiglini2024evidence}
Famiglini, L., et~al.: Evidence-based xai: An empirical approach to design more effective and explainable decision support systems. Computers in biology and medicine  \textbf{170},  108042 (2024)

\bibitem{ge2021yoloxexceedingyoloseries}
Ge, Z., et~al.: Yolox: Exceeding yolo series in 2021 (2021)

\bibitem{10.1007/978-3-031-09037-0_8}
Gomez, T., et~al.: Metrics for saliency map evaluation of deep learning explanation methods. In: Pattern Recognition and Artificial Intelligence. pp. 84--95. Springer International Publishing, Cham (2022)

\bibitem{8689279}
Grami, A.: The Gaussian Distribution, pp. 201--238 (2019)

\bibitem{Gunning_Aha_2019}
Gunning, D., Aha, D.: Darpa’s explainable artificial intelligence (xai) program. AI Magazine  \textbf{40}(2),  44--58 (Jun 2019)

\bibitem{8237584}
He, K., et~al.: Mask r-cnn. In: 2017 IEEE International Conference on Computer Vision (ICCV). pp. 2980--2988 (2017)

\bibitem{hedstrom2023quantus}
Hedstr{\"{o}}m, A., et~al.: Quantus: An explainable ai toolkit for responsible evaluation of neural network explanations and beyond. Journal of Machine Learning Research  \textbf{24}(34),  1--11 (2023)

\bibitem{hoofnagle2019european}
Hoofnagle, C.J., et~al.: The european union general data protection regulation: what it is and what it means. Information \& Communications Technology Law  \textbf{28}(1),  65--98 (2019)

\bibitem{10297629}
Kadir, M.A., et~al.: Evaluation metrics for xai: A review, taxonomy, and practical applications. In: 2023 IEEE 27th International Conference on Intelligent Engineering Systems (INES). pp. 000111--000124 (2023)

\bibitem{KAKOGEORGIOU2021102520}
Kakogeorgiou, I., Karantzalos, K.: Evaluating explainable artificial intelligence methods for multi-label deep learning classification tasks in remote sensing. International Journal of Applied Earth Observation and Geoinformation  \textbf{103},  102520 (2021)

\bibitem{law2019cornernetdetectingobjectspaired}
Law, H., Deng, J.: Cornernet: Detecting objects as paired keypoints (2019)

\bibitem{ijcai2023p747}
Le, P.Q., et~al.: Benchmarking explainable ai - a survey on available toolkits and open challenges. In: Elkind, E. (ed.) Proceedings of the Thirty-Second International Joint Conference on Artificial Intelligence, {IJCAI-23}. pp. 6665--6673. International Joint Conferences on Artificial Intelligence Organization (8 2023), survey Track

\bibitem{10.1007/978-3-319-10602-1_48}
Lin, T.Y., et~al.: Microsoft coco: Common objects in context. In: Fleet, D., Pajdla, T., Schiele, B., Tuytelaars, T. (eds.) Computer Vision -- ECCV 2014. pp. 740--755. Springer International Publishing, Cham (2014)

\bibitem{Liu_2016}
Liu, W., et~al.: SSD: Single Shot MultiBox Detector, p. 21–37. Springer International Publishing (2016)

\bibitem{9710580}
Liu, Z., et~al.: Swin transformer: Hierarchical vision transformer using shifted windows. In: 2021 IEEE/CVF International Conference on Computer Vision (ICCV). pp. 9992--10002 (2021)

\bibitem{app12199423}
Lopes, P., et~al.: Xai systems evaluation: A review of human and computer-centred methods. Applied Sciences  \textbf{12}(19) (2022)

\bibitem{10.1145/3387166}
Mohseni, S., et~al.: A multidisciplinary survey and framework for design and evaluation of explainable ai systems. ACM Trans. Interact. Intell. Syst.  \textbf{11}(3–4) (Sep 2021)

\bibitem{Molnar2022}
Molnar, C., et~al.: General Pitfalls of Model-Agnostic Interpretation Methods for Machine Learning Models, pp. 39--68. Springer International Publishing, Cham (2022)

\bibitem{MORADI2024109183}
Moradi, M., et~al.: Model-agnostic explainable artificial intelligence for object detection in image data. Engineering Applications of Artificial Intelligence  \textbf{137},  109183 (2024)

\bibitem{mumuni2025explainableartificialintelligencexai}
Mumuni, F., Mumuni, A.: Explainable artificial intelligence (xai): from inherent explainability to large language models (2025)

\bibitem{Natarajan2024}
Natarajan, S., et~al.: Robust diagnosis and meta visualizations of plant diseases through deep neural architecture with explainable ai. Scientific Reports  \textbf{14}(1),  13695 (Jun 2024)

\bibitem{10.1145/3583558}
Nauta, M., et~al.: From anecdotal evidence to quantitative evaluation methods: A systematic review on evaluating explainable ai. ACM Comput. Surv.  \textbf{55}(13s) (Jul 2023)

\bibitem{neuwirth2022eu}
Neuwirth, R.: The eu artificial intelligence act. The EU Artificial Intelligence Act  \textbf{106} (2022)

\bibitem{nguyen2025human}
Nguyen, H., et~al.: Human-centered explainable psychiatric disorder diagnosis system using wearable ecg monitors

\bibitem{ijcai2024p1025}
Nguyen, H., et~al.: Langxai: Integrating large vision models for generating textual explanations to enhance explainability in visual perception tasks. In: Proceedings of the Thirty-Third International Joint Conference on Artificial Intelligence, {IJCAI-24}. pp. 8754--8758 (8 2024)

\bibitem{NGUYEN2025102782}
Nguyen, H.T.T., et~al.: Xedgeai: A human-centered industrial inspection framework with data-centric explainable edge ai approach. Information Fusion  \textbf{116},  102782 (2025)

\bibitem{nguyen2024efficientconciseexplanationsobject}
Nguyen, K., et~al.: Efficient and {Concise} {Explanations} for {Object} {Detection} with {Gaussian}-{Class} {Activation} {Mapping} {Explainer}. Proceedings of the Canadian Conference on Artificial Intelligence  (2024)

\bibitem{nguyen2023towards}
Nguyen, T.T.H., et~al.: Towards trust of explainable ai in thyroid nodule diagnosis. In: International Workshop on Health Intelligence. pp. 11--26. Springer (2023)

\bibitem{10444383}
Nguyen, T.T.H., et~al.: Enhancing the fairness and performance of edge cameras with explainable ai. In: 2024 IEEE International Conference on Consumer Electronics (ICCE). pp.~1--4 (2024)

\bibitem{padmanabhan2023dextdetectorexplanationtoolkit}
Padmanabhan, D.C., et~al.: Dext: Detector explanation toolkit. In: World Conference on Explainable Artificial Intelligence. pp. 433--456. Springer (2023)

\bibitem{petsiuk2018riserandomizedinputsampling}
Petsiuk, V., Das, A., Saenko, K.: Rise: Randomized input sampling for explanation of black-box models. In: Proceedings of the British Machine Vision Conference (BMVC) (2018)

\bibitem{Petsiuk_2021_CVPR}
Petsiuk, V., et~al.: Black-box explanation of object detectors via saliency maps. In: Proceedings of the IEEE/CVF Conference on Computer Vision and Pattern Recognition (CVPR). pp. 11443--11452 (June 2021)

\bibitem{redmon2018yolov3incrementalimprovement}
Redmon, J., Farhadi, A.: Yolov3: An incremental improvement (2018)

\bibitem{redmon2016lookonceunifiedrealtime}
Redmon, J., et~al.: You only look once: Unified, real-time object detection (2016)

\bibitem{NIPS2015_14bfa6bb}
Ren, S., et~al.: Faster r-cnn: Towards real-time object detection with region proposal networks. In: Cortes, C., et~al. (eds.) Advances in Neural Information Processing Systems. vol.~28. Curran Associates, Inc. (2015)

\bibitem{ren2016fasterrcnnrealtimeobject}
Ren, S., et~al.: Faster r-cnn: Towards real-time object detection with region proposal networks (2016)

\bibitem{10.1145/2939672.2939778}
Ribeiro, M.T., et~al.: "why should i trust you?": Explaining the predictions of any classifier. In: Proceedings of the 22nd ACM SIGKDD International Conference on Knowledge Discovery and Data Mining. p. 1135–1144. KDD '16, Association for Computing Machinery, New York, NY, USA (2016)

\bibitem{make3030033}
Sejr, J.H., et~al.: Surrogate object detection explainer (sodex) with yolov4 and lime. Machine Learning and Knowledge Extraction  \textbf{3}(3),  662--671 (2021)

\bibitem{Selvaraju_2017_ICCV}
Selvaraju, R.R., et~al.: Grad-cam: Visual explanations from deep networks via gradient-based localization. In: Proceedings of the IEEE International Conference on Computer Vision (ICCV) (Oct 2017)

\bibitem{inproceedings}
Song, L., et~al.: Explainable artificial intelligence to interpret spatially-explicit impacts of future climate change on species distribution (01 2024)

\bibitem{s24216776}
Tahir, H.A., et~al.: A novel hybrid xai solution for autonomous vehicles: Real-time interpretability through lime–shap integration. Sensors  \textbf{24}(21) (2024)

\bibitem{abc}
Tarmissi, K., et~al.: Explainable artificial intelligence with fusion-based transfer learning on adverse weather conditions detection using complex data for autonomous vehicles. AIMS Mathematics  \textbf{9}(12),  35678--35701 (2024)

\bibitem{Teng2022}
Teng, Q., et~al.: A survey on the interpretability of deep learning in medical diagnosis. Multimedia Systems  \textbf{28}(6),  2335--2355 (Dec 2022)

\bibitem{9010746}
Tian, Z., et~al.: Fcos: Fully convolutional one-stage object detection. In: 2019 IEEE/CVF International Conference on Computer Vision (ICCV). pp. 9626--9635 (2019)

\bibitem{pmlr-v222-truong24a}
Truong, V.B., et~al.: Towards better explanations for object detection. In: Yanıkoğlu, B., Buntine, W. (eds.) Proceedings of the 15th Asian Conference on Machine Learning. Proceedings of Machine Learning Research, vol.~222, pp. 1385--1400. PMLR (11--14 Nov 2024)

\bibitem{wang2020score}
Wang, H., et~al.: Score-cam: Score-weighted visual explanations for convolutional neural networks. In: Proceedings of the IEEE/CVF conference on computer vision and pattern recognition workshops. pp. 24--25 (2020)

\bibitem{WOOD2022102391}
Wood, D.A., et~al.: Deep learning models for triaging hospital head mri examinations. Medical Image Analysis  \textbf{78},  102391 (2022)

\bibitem{Yamauchi_2024_CVPR}
Yamauchi, T.: Spatial sensitive grad-cam++: Improved visual explanation for object detectors via weighted combination of gradient map. In: Proceedings of the IEEE/CVF Conference on Computer Vision and Pattern Recognition (CVPR) Workshops. pp. 8164--8168 (June 2024)

\bibitem{9897350}
Yamauchi, T., Ishikawa, M.: Spatial sensitive grad-cam: Visual explanations for object detection by incorporating spatial sensitivity. In: 2022 IEEE International Conference on Image Processing (ICIP). pp. 256--260 (2022)

\bibitem{YANG202229}
Yang, G., et~al.: Unbox the black-box for the medical explainable ai via multi-modal and multi-centre data fusion: A mini-review, two showcases and beyond. Information Fusion  \textbf{77},  29--52 (2022)

\bibitem{zhang2018top}
Zhang, J., et~al.: Top-down neural attention by excitation backprop. International Journal of Computer Vision  \textbf{126}(10),  1084--1102 (2018)

\bibitem{zhang2021group}
Zhang, Q., et~al.: Group-cam: Group score-weighted visual explanations for deep convolutional networks. arXiv preprint arXiv:2103.13859  (2021)

\bibitem{10478163}
Zhao, C., et~al.: { Gradient-Based Instance-Specific Visual Explanations for Object Specification and Object Discrimination }. IEEE Transactions on Pattern Analysis \& Machine Intelligence  \textbf{46}(09),  5967--5985 (Sep 2024)

\end{thebibliography}

\end{document}